# NavFormer: A Transformer Architecture for Robot Target-Driven Navigation in Unknown and Dynamic Environments

Haitong Wang, Aaron Hao Tan, *Student Member, IEEE* and Goldie Nejat, *Member, IEEE*

*Abstract*—In unknown cluttered and dynamic environments such as disaster scenes, mobile robots need to perform target-driven navigation in order to find people or objects of interest, where the only information provided about these targets are images of the individual targets. In this paper, we introduce NavFormer, a novel end-to-end transformer architecture developed for robot target-driven navigation in unknown and dynamic environments. NavFormer leverages the strengths of both 1) transformers for sequential data processing and 2) self-supervised learning (SSL) for visual representation to reason about spatial layouts and to perform collision-avoidance in dynamic settings. The architecture uniquely combines dual-visual encoders consisting of a static encoder for extracting invariant environment features for spatial reasoning, and a general encoder for dynamic obstacle avoidance. The primary robot navigation task is decomposed into two sub-tasks for training: single robot exploration and multi-robot collision avoidance. We perform cross-task training to enable the transfer of learned skills to the complex primary navigation task. Simulated experiments demonstrate that NavFormer can effectively navigate a mobile robot in diverse unknown environments, outperforming existing state-of-the-art methods. A comprehensive ablation study is performed to evaluate the impact of the main design choices of NavFormer. Furthermore, real-world experiments validate the generalizability of NavFormer.

## I. INTRODUCTION

Mobile robots can be used to search for potential victims in unknown environments including in urban disaster environments [1], [2], in buildings engulfed by fire [3], and/or unstructured outdoor environments [4]. Images of potential victims can be provided to the robots for them to search a disaster environment for these specific individuals, while avoiding collisions with rescue workers, victims, and other robots. In this paper, we address the problem of robot target-driven navigation (TDN) in unknown and dynamic environments. This problem requires a mobile robot to navigate an unknown and dynamic environment using only an onboard RGB camera to search for a static target given its image.

TDN in unknown and dynamic environments is a challenging problem as: 1) there are no global maps of the environment available, therefore, a robot needs to reason about the spatial layout of the environment based on its own partial observations to prevent deadlocks and redundant coverage [5], and 2) the presence of dynamic obstacles needs to be detected for collision avoidance [6] and for spatial reasoning of the environment [7]. In this paper, we assume that for each robot, the dynamic obstacles are other moving robots.

To-date, existing robot TDN methods for unknown environments have mainly used: 1) deep reinforcement learning (DRL) [5], [8]-[12] or 2) imitation learning (IL) [13]. Images of targets have consisted of either indoor scenes (e.g., kitchen, bedroom) [8] or household objects (e.g., chair, microwave) [10]. These methods take RGB images as observations and the target image as input into a convolutional neural network (CNN) to extract features (e.g., geometry, patterns) and encode them into a latent vector. Namely, for DRL methods, navigation actions are generated using either fully connected layers (FCL) [8], long short term memory (LSTM) [10], or attention-based memory retrieval [5] approaches. For IL methods [13], the robot action is generated by predicting the next expected observation (NEO) using an inverse dynamics model. These aforementioned TDN methods have mainly been applied to static environments. However, in dynamic environments, they may result in degraded performance due to the presence of moving obstacles that are treated as static. This may lead to misinterpretation of the spatial layout of the environment [7], which in turn can result in ineffective navigation decisions.

In this paper, we propose NavFormer, a novel end-to-end DL architecture consisting of a dual-visual encoder module and a transformer-based navigation network to address *for the first time* the problem of TDN in unknown and dynamic environments. NavFormer utilizes a decoder-only transformer [14] to make navigation decisions conditioned on both the target image and robot trajectory history. To obtain high-quality datasets for navigation policy learning, we decompose the task of TDN in dynamic environments into two well-studied sub-tasks in the literature: single-robot exploration, and multi-robot collision avoidance. Our main contributions in this paper include: 1) the development of the first end-to-end DL approach for robot target-driven navigation in unknown dynamic environments; 2) the incorporation of a dual-visual encoder system to extract static and general (i.e., static and dynamic) features for reasoning the spatial layouts of environments and collision avoidance, which is trained using self-supervised learning (SSL); and 3) the development of a cross-task training

Manuscript received February 9, 2024; Revised May 8, 2024; Accepted June 3, 2024. This paper was recommended for publication by Editor Pascal Vasseur upon evaluation of the Associate Editor and Reviewers' comments. This work was supported in part by the Natural Sciences and Engineering Research Council of Canada (NSERC), and in part by the Canada Research Chairs program (CRC). *(Corresponding author: Haitong Wang.)*

The authors are with the Autonomous Systems and Biomechatronics Laboratory (ASBLab), Department of Mechanical and Industrial Engineering, University of Toronto, Toronto, ON M5S 3G8, Canada (e-mail: haitong.wang@mail.utoronto.ca;aaronhao.tan@utoronto.ca; nejat@mie.utoronto.ca).

Digital Object Identifier (DOI): see top of this page.

strategy to train NavFormer on the two subtasks of single robot exploration and multi-robot collision avoidance.

## II. RELATED WORKS

Herein, we discuss the pertinent literature on: 1) robot target-driven navigation in static unknown environments, 2) robot navigation in dynamic environments, and 3) cross-task training.

### A. Robot Target-Driven Navigation in Static Environments

Previous work has mainly used: 1) DRL [5], [8]-[12] or 2) IL [13] to address TDN tasks. These methods used Siamese CNN such as ResNet [15] to generate a joint embedding from robot image observations and the target image, and then they used this joint embedding to generate navigation actions. In [8], the Asynchronous Advantage Actor Critic (A3C) [16] was used to train scene-specific FCLs to generate navigation actions using joint embedding. Each environment required training on a set of scene-specific layers. In [9], object recognition was used to generate bounding boxes of the target object via two FCLs, and then used by an action prediction module to generate robot navigation actions. In [10], the Importance Weighted Actor-Learner Architecture (IMPALA) [18] was used to train a navigation network with LSTM [17] to account for historical observations for robot navigation in 3D mazes. In [5], a memory buffer and an attention module were used to further improve a robot's temporal reasoning via long-term memory for indoor navigation.

In [11], a DRL architecture combined with semantic scene priors was presented. It was constructed Using a knowledge graph to represent semantic relationships between objects in the scene. An actor-critic network was used to generate actions. In [12], hierarchical policy learning with Intrinsic-Extrinsic Modeling (HIEM) was developed by using a high-level policy that generated sub-goals to guide target search and a low-level policy to generate navigation actions.

In [13], a generative IL module was used to predict the NEO, which was then used by an action prediction module to predict the robot's action.

### B. Robot Navigation in Dynamic Environments

Robot navigation methods in dynamic environments can be categorized as: 1) classical methods [6], [19], [20], or 2) learning-based methods [21]-[25]. Learning-based methods include: 1) DRL methods [21]-[23], and 2) hybrid methods [24], [25] using both DRL and IL.

*1) Classical Methods:* In [6], the Velocity Obstacle (VO) method was used for robot collision avoidance in 2D dynamic environments by generating a potential collision area based on the velocities, positions and sizes of a robot and its nearby moving obstacle (another robot), and then selecting a robot velocity that avoided this area. Reciprocal VO (RVO) [19] address issues in movement oscillation of VO and the Non-Holonomic Optimal Reciprocal Collision Avoidance (NH-ORCA) approach [20] extend VO methods for robots with non-holonomic constraints.

*2) Learning-based Methods:* In [21], a multi-robot collision avoidance architecture was developed by using a CNN to directly map laser scans to robot velocities. In [22], a Gated Recurrent Unit (GRU) [27] was incorporated in [21] to account for historical observations to improve temporal reasoning in unknown environments. In [23], RL-RVO used a set of sequential VO and RVO vectors representing the states of nearby obstacles and a Bidirectional GRU network to generate navigation velocity.

In [24], a Hybrid CPU/GPU A3C for Collision Avoidance with DRL (GA3C-CADRL) method used a LSTM to encode spatial information of nearby obstacles and was trained using IL and then DRL. In [25], the Pathfinding via Reinforcement and Imitation Multi-Agent Learning (PRIMAL) was developed for multi-robot navigation by using a CNN to encode the local 2D map and a LSTM to generate robot actions. During training, PRIMAL randomly switched between DRL and IL to learn a navigation policy that improved navigation performance.

### C. Cross-task Training

Cross-task training considers training an agent on multiple tasks ranging from video games [18],[28] to robotic applications such as tracking [29] and navigation [10]. For example, in [18], an off-policy actor-critic architecture, IMPALA, was developed to train an RL agent to complete multiple tasks with decoupled acting and learning and off-policy correction. In [29], an End-to-End Visual Active Tracking (E-VAT) method divided the VAT task into two sequential sub-tasks: exploration and tracking. The sub-tasks were trained concurrently using an asymmetric actor-critic architecture and IMPALA. In [10], the TDN task in static environments was addressed. Training was split into localization and navigation phases. The localization network was trained by self-supervised learning for target object localization, and the navigation policy was trained via IMPALA for target navigation. In [28], a Multi-Game Decision Transformer was utilized to train a single transformer model to play multiple video games by learning from a diverse offline dataset with both expert and non-expert data.

### D. Summary of Limitations

The aforementioned TDN methods [5], [8]-[13] address the problem of a single robot navigating to a target location in an unknown static environment. However, these methods do not consider dynamic obstacles (i.e., other robots) in their environments, leading to inaccuracies in spatial reasoning and degraded navigation performance [7]. Robot navigation in dynamic environments has been achieved using classical methods [6], [19], [20], or learning-based methods [21]-[25]. These methods consist of only local planning schemes without considering global spatial layouts, resulting in robots that become trapped in local minima (e.g., dead ends) [31]. Furthermore, they cannot find a target provided in an RGB image without a given location. To address these limitations, we have developed NavFormer, the first DL method for robot TDN in unknown and dynamic environments. Our method utilizes cross-task training on a decision transformer architecture [32] developed for the TDN task, encompassing both exploration and multi-robot collision avoidance tasks.

## III. TARGET-DRIVEN NAVIGATION PROBLEM IN UNKNOWN AND DYNAMIC ENVIRONMENTS

### A. Problem Definition

Robot target-driven navigation in unknown and dynamic environments describes the following problem: a mobile robot

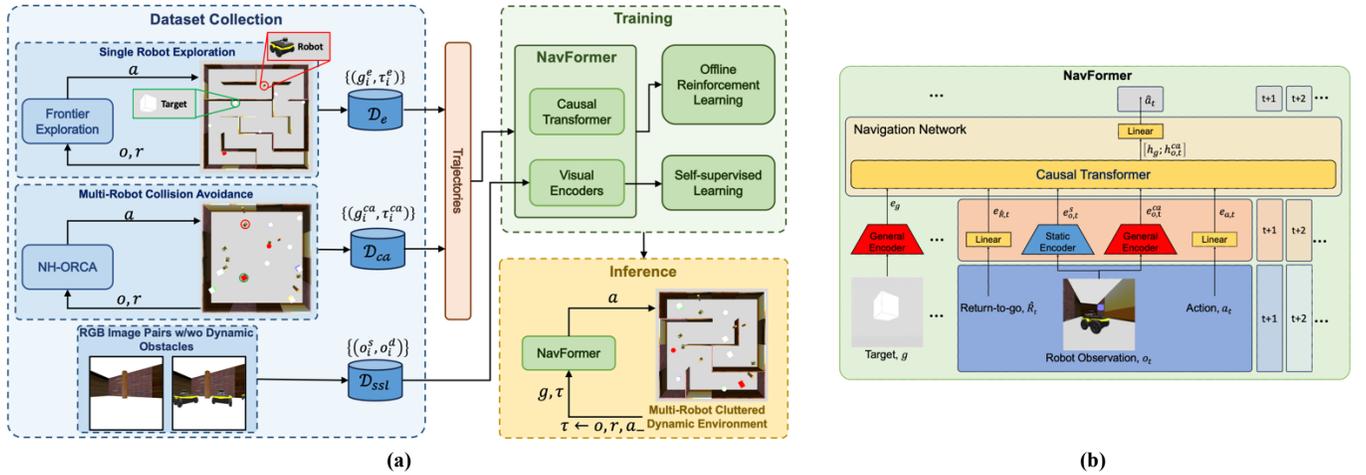

Fig. 1. (a) NavFormer TDN architecture consisting of Dataset Collection, Training and Inference subsystems. (b) NavFormer structure consisting of multi-modal input sequence, dual-visual encoders, and navigation network.

$r$ needs to navigate to a target $I$ utilizing only an RGB image of the target $g$ and visual observations $o \in \Omega$ (i.e., RGB images) of the environment obtained from an onboard camera. The dynamic obstacles (i.e., other robots) in the environment are represented by a set $M$. There are no a priori global maps of the environment available and the 2D location of the static target object $l_I$ is unknown. The target object is defined by a 3D geometric shape $b \in B$ and color $c \in C$. The objective of the robot $r$ is to minimize the expected travel distance $d$ between the robot's start location $l_s$ and target location $l_I$:

$$\min \mathrm{E}\left[d(l_s, l_I)\right]. \quad (1)$$

*B. GC-POMDP for Robot Target-Driven Navigation*

We model the robot TDN problem as a goal-conditioned partially observable Markov decision process (GC-POMDP). GC-POMDP is described as a tuple $(\mathcal{S}, \mathcal{G}, \mathcal{A}, \mathcal{P}, \mathcal{R}, \Omega, \mathcal{O})$, where $\mathcal{S}$ denotes the state space, and $\mathcal{G}$ is the set of target RGB images. Robot actions, $a \in \mathcal{A}$, are represented by a 2D vector of linear and angular velocity. $\mathcal{P}$ is the state transition function $\mathcal{P}(s, a, s') = p(s'|s, a)$. $\mathcal{R}$ is the reward function, $r = \mathcal{R}(s, a)$. $\Omega$ is the observation space and $\mathcal{O}$ is the observation probability function $\mathcal{O}(s', a, o) = p(o|s', a)$. At each time step, the robot observes the environment, takes an action, then transitions to the next state and receives a reward.

The objective is to learn a policy $\pi_\theta(a|g, \tau)$ that is conditioned on the target image $g$ and robot historical trajectory $\tau$ to maximize the expected return: $E[\sum_{t=1}^T r_t]$. The robot historical trajectory $\tau$ consists of returns-to-go $\hat{R}$, observations $o$, and actions $a$:

$$\tau_t = (\hat{R}_1, o_1, a_1, \hat{R}_2, o_2, a_2, \ldots, \hat{R}_t, o_t), \quad (2)$$

where the return-to-go, $\hat{R}_t$, is defined as the desired total sum of rewards to achieve from the current timestep $t$ to the terminal timestep $T$ of the episode [32]:

$$\hat{R}_t = \sum_{k=t}^T r_k. \quad (3)$$

Return-to-go is used to generate actions conditioned on desired returns rather than past rewards. The return-to-go at the first timestep, $\hat{R}_1$, is a user specified desired total rewards.

## IV. NAVFORMER ARCHITECTURE

The proposed NavFormer TDN architecture, Fig. 1(a), consists of three subsystems: 1) Dataset collection: to obtain datasets containing robot trajectories for policy learning, and RGB image pairs for representation learning; 2) Training: to train both the NavFormer model using offline reinforcement learning and the dual-visual encoders (DVE) using self-supervised learning, and 3) Inference: which uses the trained NavFormer model for the TDN task in unknown dynamic environments.

In this section, we will discuss the development of the NavFormer structure, Fig. 1(b). NavFormer contains: 1) a Multi-modal Input Sequence containing the target image $g$ and robot trajectory $\tau$, 2) a Dual-visual encoder module that separately extracts static and general (i.e., static and dynamic) features from visual observations, and 3) a transformer-based Navigation Network (NavN) that is conditioned on the multi-modal input embeddings to generate navigation actions.

*A. Multi-Modal Input Sequence*

The multi-modal input sequence $s_t$ consists of the RGB image of the target $g$ and robot trajectory $\tau_t$:

$$s_t = (g, \hat{R}_1, o_1, a_1, \hat{R}_2, o_2, a_2, \ldots, \hat{R}_t, o_t). \quad (4)$$

The sequence is converted to embeddings of the same dimension $d_e = 128$. We use a linear layer to project $\hat{R}_i$ and $a_i$ to embeddings $e_{\hat{R},i}$ and $e_{a,i}$, respectively. These embeddings are combined with the embeddings output by the DVE and provided to the NavN.

*B. Dual-Visual Encoders*

Our architecture utilizes DVE for effective TDN: 1) the static encoder $f_s$ extracts spatial features for environmental layout reasoning, which is essential for exploring the unknown environment, and 2) the general encoder $f_g$ extracts obstacle-specific features from nearby static and dynamic obstacles, facilitating collision avoidance during navigation. Two separate encoders are used to explicitly extract features specific to the two sub-tasks. Task-specific feature extraction has been shown to enhance policy learning in robot navigation [33].

Each encoder consists of a CNN with three convolutional layers with a kernel size, stride, and output channel of (8, 4, 32), (4, 2, 64), (3, 2, 64) [34]. All images (i.e., $g$ and $o$) are in the dimension of (84, 84, 3). Given an input sequence $s_t$, $g$ is used by $f_g$ to generate a target embedding $e_g$. Furthermore, each observation $o_i$ is used by $f_s$ and $f_g$ to generate a static embedding $e_{o,i}^{s}$ and a general embedding $e_{o,i}^{ca}$, respectively. $e_{o,i}^{s}$ and $e_{o,i}^{ca}$ are then combined with the embedding of returns-to-go $e_{\hat{R},i}$ and actions $e_{a,i}$ from Section IV.A. Thus, $s_t$ is converted into a sequence of embeddings $E_{s,t}$ which is then used by the NavN to generate navigation actions:

$$E_{s,t} = (e_g, e_{\hat{R},1}, e_{o,1}^{s}, e_{o,1}^{ca}, e_{a,1}, \ldots, e_{\hat{R},t}, e_{o,t}^{s}, e_{o,t}^{ca}). \quad (5)$$

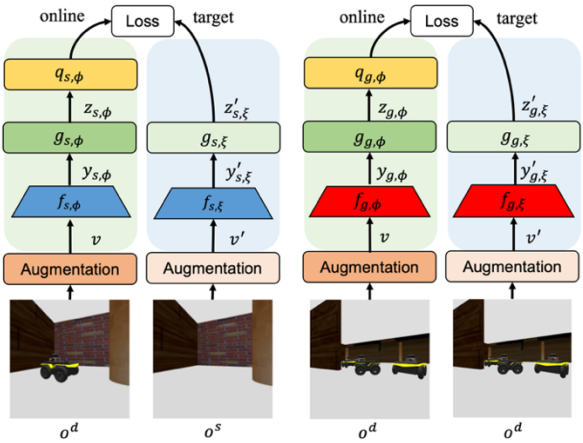

**Fig. 2.** BYOL procedure for the static and general encoder.

*1) Training Loss:* In order to train the two visual encoders, the Bootstrap Your Own Latent (BYOL) [35] self-supervised learning method is used. For each encoder, BYOL uses two neural networks, Fig. 2: 1) an online network parameterized by $\phi$ that consists of an encoder ($f_{s,\phi}, f_{g,\phi}$), a projector ($g_{s,\phi}, g_{g,\phi}$), and a predictor ($q_{s,\phi}, q_{g,\phi}$); and 2) a target network parameterized by $\xi$ that consists of an encoder ($f_{s,\xi}, f_{g,\xi}$), and a projector ($g_{s,\xi}, g_{g,\xi}$). Herein, we represent these parameters in general as ($f_\phi, g_\phi, q_\phi, f_\xi, g_\xi$) as the training of $f_s$ and $f_g$ follow the same procedure. The online network and the target network share the same architecture for the encoder and the projector. The projector and the predictor are represented by a multiple-layer perceptron (MLP). The weights of the target network are an exponential moving average of the online network weights.

Two distributions of image augmentations including random resizing, cropping, color jittering were applied to obtain two augmented views $v$ and $v'$ from the input observation image $o$. The online network takes $v$ as input, and outputs representation $y_\phi = f_\phi(v)$, projection $z_\phi = g_\phi(y_\phi)$, and a prediction $q_\phi(z_\phi)$. The target network takes as input $v'$, and outputs representation $y'_\xi = f_\xi(v')$ and projection $z'_\xi = g_\xi(y'_\xi)$. The loss function is defined as the mean squared error (MSE) between the normalized online prediction and target projection:

$$\mathcal{L}_{\phi,\xi} = \left\| \frac{q_\phi(z_\phi)}{\|q_\phi(z_\phi)\|_2} - \frac{z'_\xi}{\|z'_\xi\|_2} \right\|_2^2. \quad (6)$$

BYOL uses a symmetrical loss by swapping $v$ and $v'$ as input to the online and target networks again to compute $\tilde{\mathcal{L}}_{\phi,\xi}$:

$$\mathcal{L}_{\phi,\xi}^{\text{BYOL}} = \mathcal{L}_{\phi,\xi} + \tilde{\mathcal{L}}_{\phi,\xi}. \quad (7)$$

The final BYOL loss is the summation of the loss for $f_s$, $\mathcal{L}_{s,\phi,\xi}^{\text{BYOL}}$ and the loss for $f_g$, $\mathcal{L}_{g,\phi,\xi}^{\text{BYOL}}$:

$$\mathcal{L}^{\text{BYOL}} = \mathcal{L}_{s,\phi,\xi}^{\text{BYOL}} + \mathcal{L}_{g,\phi,\xi}^{\text{BYOL}}. \quad (8)$$

To ensure that the static encoder learns only the static features from the environment, we modified the standard BYOL procedure [35] by generating the two augmented views ($v, v'$) from two different images ($o^s, o^d$) of the same view of the scene. $o^d$ denotes an RGB image that contains dynamic obstacles (i.e., other robots) while $o^s$ denotes an image that contains only static obstacles. Therefore, the static encoder learns only the common static obstacle features shared between the images ($o^s, o^d$). For the general encoder, we follow the standard BYOL procedure [42] and generate two different augmented views from the same image $o^d$.

*C. Navigation Network*

The NavN is used to generate the robot's action $a_{t+1}$ given the sequence of embeddings $E_{s,t}$. Our NavN uses a causal transformer (CT) model based on the Generative Pre-trained Transformer 2 (GPT-2) [14]. We use CT as its self-attention mechanism can effectively utilize sequential memory of past static embeddings $e_o^s$, enabling the network to reason about the spatial layout of the environment based on the robot trajectory $\tau_t$. The CT can also implicitly infer the motion of dynamic obstacles by dynamically adjusting the weight of each general embedding $e_o^{ca}$ from consecutive image observations via self-attention layers.

The structure of our NavN consists of three stacked transformer blocks, and a linear layer. To understand the temporal and multi-modal nature of the input data, positional encodings are introduced as learnable embeddings for each timestep and modality, which are added to $E_{s,t}$.

Dropout is applied to the $E_s$ to prevent overfitting [36]. $E_s$ is then sent into $N = 3$ stacked transformer blocks, each consisting of a masked multi-head attention (MMHA) submodule and a MLP submodule. The context length of each transformer block is 1024. In the MMHA submodule, the embeddings $E_s$ are linearly projected into $h = 2$ heads of keys $K$, queries $Q$ and values $V$ in the dimension of $d_k = 64$ for self-attention. The MMHA submodule is connected to a MLP that has a hidden layer and residual connection. The last transformer block outputs a sequence of hidden states $H_s$:

$$H_s = (h_g, h_{\hat{R},1}, h_{o,1}^{s}, h_{o,1}^{ca}, \ldots, h_{\hat{R},t}, h_{o,t}^{s}, h_{o,t}^{ca}). \quad (9)$$

We concatenate the target hidden state $h_g$ with the observation hidden state $h_{o,i}^{ca}$ and provide $[h_g; h_{o,t}^{ca}]$ into a linear layer to predict the next navigation action $\hat{a}_i$. This hidden state concatenation helps the CT to implicitly detect the target; thereby, improving the navigation policy.

*1) Training Loss:* We utilize the Decision Transformer [32] offline RL method for training the NavN. The loss of the NavN is computed using the MSE between the predicted actions $\hat{a}_i$ and ground-truth actions $a_i$:

$$\mathcal{L}^{DT} = \frac{1}{T}\sum_{t=1}^{T} \| a_t - \hat{a}_t \|^2. \quad (10)$$

V. DATASET COLLECTION

We collected three datasets in 3D simulated environments from mobile robots to train NavFormer, Fig. 1(a): 1) Robot Exploration Dataset, $\mathcal{D}_e$, 2) Collision Avoidance Dataset, $\mathcal{D}_{ca}$, and 3) Representation Learning Dataset, $\mathcal{D}_{ssl}$. The target objects include various geometric shapes (sphere, box) and colors (green, red, white, yellow, blue).

**Exploration Dataset $\mathcal{D}_e$:** This dataset comprises 3,496 robot trajectories and the corresponding target RGB images, totaling 342,954 timesteps for a single robot exploration task in 3D environment with only static obstacles (i.e., walls). Each robot trajectory was obtained in a different random environment with size ranging from 5m × 5m to 17.5m × 17.5m. The trajectory $\tau$ consists of returns-to-go $\hat{R}$, observations $o$, and actions $a$. To generate this dataset, the Frontier Exploration method in [38] was used to select the nearest frontier location for the robot. The robot planned a global path using A* and followed this path using DWA [30] as the local planner. $\mathcal{D}_e$ is used to train NavFormer to learn the skill of exploration.

TABLE I
COMPARISON BETWEEN NAVFORMER AND SOTA METHODS FOR UNSEEN MAZE ENVIRONMENTS

| Setup | | VRW | | | DSAC | | | MA-TDN | | | DT | | | FCA | | | NavFormer (ours) | | |
|---|---|---|---|---|---|---|---|---|---|---|---|---|---|---|---|---|---|---|---|
| Env size (m) | # of robots | SR | SPL | CR | SR | SPL | CR | SR | SPL | CR | SR | SPL | CR | SR | SPL | CR | SR | SPL | CR |
| 7.5 × 7.5 | 2 | 39.17% | 0.287 | 0.691 | 53.33% | 0.377 | 0.316 | 61.67% | 0.315 | 0.314 | 57.50% | 0.419 | 0.405 | 94.17% | 0.653 | 0.044 | 80.83% | 0.562 | 0.228 |
| | 4 | 33.33% | 0.220 | 0.694 | 47.08% | 0.353 | 0.395 | 46.67% | 0.248 | 0.542 | 47.92% | 0.395 | 0.529 | 80.83% | 0.598 | 0.142 | 66.25% | 0.465 | 0.325 |
| | 6 | 32.78% | 0.187 | 0.711 | 42.78% | 0.337 | 0.493 | 43.06% | 0.234 | 0.559 | 40.28% | 0.332 | 0.608 | 76.11% | 0.580 | 0.101 | 54.44% | 0.435 | 0.430 |
| 12.5 × 12.5 | 2 | 16.67% | 0.100 | 0.678 | 24.17% | 0.179 | 0.324 | 30.00% | 0.159 | 0.387 | 39.17% | 0.258 | 0.372 | 85.00% | 0.599 | 0.070 | 53.33% | 0.374 | 0.251 |
| | 4 | 15.83% | 0.104 | 0.680 | 19.17% | 0.133 | 0.460 | 23.33% | 0.137 | 0.394 | 26.25% | 0.201 | 0.430 | 73.33% | 0.482 | 0.096 | 48.89% | 0.378 | 0.223 |
| | 6 | 13.61% | 0.101 | 0.716 | 23.89% | 0.18 | 0.550 | 23.06% | 0.118 | 0.488 | 31.94% | 0.231 | 0.530 | 60.28% | 0.454 | 0.071 | 48.75% | 0.359 | 0.355 |
| 17.5 × 17.5 | 2 | 5.83% | 0.053 | 0.666 | 10.00% | 0.084 | 0.371 | 20.83% | 0.117 | 0.342 | 34.17% | 0.222 | 0.352 | 57.50% | 0.395 | 0.068 | 44.17% | 0.316 | 0.279 |
| | 4 | 5.00% | 0.032 | 0.622 | 10.00% | 0.085 | 0.472 | 13.75% | 0.078 | 0.494 | 22.08% | 0.151 | 0.449 | 47.50% | 0.331 | 0.100 | 38.33% | 0.279 | 0.204 |
| | 6 | 5.83% | 0.032 | 0.585 | 10.56% | 0.092 | 0.538 | 16.39% | 0.098 | 0.554 | 22.50% | 0.169 | 0.546 | 40.83% | 0.296 | 0.081 | 35.56% | 0.272 | 0.350 |

**Collision Avoidance Dataset $\mathcal{D}_{ca}$:** This dataset comprises 7,467 robot trajectories collected using NH-ORCA [20] and the corresponding target images, totaling 469,378 timesteps for the task of multi-robot collision avoidance in dynamic environments. Each trajectory was obtained in a 12.5 m × 12.5 m environment with the number of dynamic obstacles ranging from 2 to 8. $\mathcal{D}_{ca}$ was used to learn collision avoidance.

**Representation Learning Dataset $\mathcal{D}_{ssl}$:** This dataset, consisting of 98,000 image pairs taken in 3D mazes with dynamic obstacles, was collected across 1,000 randomly generated environments, varying in size from 5 m × 5 m to 17.5 m × 17.5 m and containing 4 to 25 dynamic obstacles. Each image pair contained two RGB images $(o_i^s, o_i^d)$ taken by the robot's onboard camera. To collect this dataset: 1) we randomly placed all dynamic obstacles in different locations, and recorded their observations (i.e., $o_i^d$) and locations; and 2) a single robot was placed at all previous locations and its own observations were captured (i.e., $o_i^s$). We aligned each $o_i^s$ with its corresponding $o_i^d$ based on the locations where $o_i^s$ and $o_i^d$ were taken. $\mathcal{D}_{ssl}$ was used for the representation learning.

## VI. TRAINING

We used a cross-task training approach for the policy learning of NavFormer. Namely, we decomposed the robot TDN task into the two sub-tasks of: 1) single robot exploration, and 2) multi-robot collision avoidance. NavFormer was trained on $\mathcal{D}_e$ and $\mathcal{D}_{ca}$ to learn exploration and collision avoidance skills. To compute $\mathcal{L}^{DT}$ (10), we alternated between $\mathcal{D}_e$ and $\mathcal{D}_{ca}$ during each training iteration to sample trajectories. To train the DVE, we use $\mathcal{D}_{ssl}$ to sample image pairs to compute $\mathcal{L}^{BYOL}$, (8). The final loss is:

$$\mathcal{L} = \mathcal{L}^{BYOL} + \mathcal{L}^{DT}. \quad (11)$$

Training was performed using an RTX3070 GPU, an AMD Ryzen Threadripper 3960X CPU and 128GB of memory. The training of the DVE utilized a batch size of 512. The Adam optimizer [39] used 0.0004 for the learning rate and weight decay. NavFormer was trained with a batch size of 25 and 150 for trajectory sampling from $\mathcal{D}_e$ and $\mathcal{D}_{ca}$, respectively. We used different batch sizes for trajectory sampling to balance the total number of frames in each batch as trajectories in $\mathcal{D}_e$ were longer than trajectories in $\mathcal{D}_{ca}$. Dropout of 0.1 is applied to the CT. The AdamW optimizer [40] was used to train the NavFormer with a learning rate of 0.0002 and weight decay of 0.0001, over a span of 10,000 iterations.

## VII. SIMULATED EXPERIMENTS

We conducted two sets of experiments to evaluate the overall performance of our NavFormer architecture: 1) a comparison study with our approach and state-of-the-art (SOTA) learning methods to compare TDN strategies on: i) unseen 3D maze environments, ii) unseen photorealistic 3D environments, and iii) unseen dynamic obstacles; and 2) an ablation study to investigate the contributions of the design choices of NavFormer.

### A. Comparison Study in Unseen Maze Environments

We used three performance metrics for these experiments: 1) mean success rate (SR) of robots reaching target objects, 2) Success weighted by normalized inverse Path Length (SPL) which measures the efficiency of the navigation method [41]:

$$\text{SPL} = \frac{1}{N_\tau} \sum_{i=1}^{N_\tau} S_i \frac{\ell_i}{\max(\wp_i, \ell_i)}, \quad (12)$$

where $N_\tau$ is the number of robot trials, $S_i$ is a binary indicator of success in trial $i$. $\ell_i$ is the shortest path length from the start location of the robot to the target location, and $\wp_i$ is the actual robot path length, and 3) mean collision rate (CR), $\text{CR} = \frac{1}{N_\tau} \sum_{i=1}^{N_\tau} \frac{N_{c,i}}{N_{e,i}}$, where $N_{c,i}$ is the number of collisions in trial $i$, $N_{e,i}$ denotes the total number of timesteps in trial $i$.

We randomly generated a total of 54 new 3D environments in Gazebo consisting of static obstacles, target objects, and dynamic obstacles (robots). For all the methods, the test environments were unseen during training. The sizes of these environments were 7.5 m × 7.5 m, 12.5 m × 12.5 m, and 17.5 m × 17.5 m, Fig. 3. Each robot had a different target object to navigate to. The target objects were randomly generated with different geometric shapes and colors, and locations. We deployed 2, 4 and 6 Jackal robots in each environment.

*1) Comparison Methods:* We benchmarked our NavFormer method against the following comparison methods:

**Velocity Random Walk (VRW)**: The velocity random walk policy randomly samples robot actions from a uniform distribution. VRW is chosen as a lower bound approach.

**Deep Siamese Actor-Critic (DSAC) [8]:** The Deep Siamese Actor-Critic model consists of Siamese CNN layers and scene-specific FCLs. We use the procedure in [13] to adapt DSAC to use a single set of scene-specific layers across all training environments in order to generalize to unseen environments.

TABLE II
COMPARISON BETWEEN NAVFORMER AND SOTA METHODS FOR UNSEEN PHOTOREALISTIC ENVIRONMENTS

|     | VRW    | DSAC   | MA-TDN | DT     | FCA    | NavFormer |
|-----|--------|--------|--------|--------|--------|-----------|
| SR  | 8.89%  | 11.11% | 17.78% | 22.22% | 82.22% | 36.67%    |
| SPL | 0.062  | 0.081  | 0.122  | 0.148  | 0.612  | 0.269     |
| CR  | 0.612  | 0.620  | 0.590  | 0.510  | 0.092  | 0.412     |

**Memory-Augmented Target-Driven Navigation (MA-TDN) [5]:** MA-TDN uses an attention module to retrieve the memory of past robot observations to generate navigation actions.

**Decision Transformer (DT) [32]:** The decision transformer uses only one visual encoder for feature extraction from visual observations. The navigation action is generated using the hidden state of the observation embedding.

**Frontier + NH-ORCA (FCA):** FCA uses frontier exploration as a high-level planner to explore the environment and uses NH-ORCA as local controller to avoid collisions with other obstacles. This method is chosen as an upper bound approach as it uses extra information unavailable to NavFormer and other SOTA learning methods, including point clouds and map, other robots' states (i.e., velocities, positions, sizes), and specific target locations. It should be noted that as FCA utilizes the above additional information, it does not directly address the TDN problem in dynamic environments using only visual observations as defined in this paper.

For both DSAC and MA-TDN, we convert their discrete action space to continuous space by outputting a mean and variance to define a Gaussian distribution for action sampling. Training of DSAC and MA-TDN were achieved with PPO [26] using the same 3D maze training environments and number of robots as NavFormer. DT was trained on $\mathcal{D}_e$ and $\mathcal{D}_{ca}$ using Eq. (10). For DT and NavFormer, $\hat{R}_1$ is initialized to be 1. The robot only receives a positive reward of 1 when it successfully navigates to the target object.

*2) Experimental Procedure:* At the beginning of each trial, robots were randomly located in the environment and assigned an RGB image of a target $g$. A robot found a target when the distance between the target and robot was within 1.5m. A trial terminated when either all robots found their corresponding targets or the total timesteps exceeded 500. Each timestep is 0.2s. A new environment was randomly generated every 10 trials. 60 trials were conducted for each experiment setup.

*3) Results:* The SR, SPL, and CR results for NavFormer and the comparison methods across the different environment sizes and number of robots are presented in Table I. In general, NavFormer consistently outperformed VRW, DSAC, MA-TDN and DT. As the environment size increased, SR and SPL decreased and CR increased for all methods due to the: 1) increased level of difficulty introduced by more static obstacles in the larger environments, and 2) longer travel distance between start and target location with the allocated time. NavFormer had higher SR and SPL and lower CR than VRW, DSAC, MA-TDN, DT across all the three environments and number of robots. NavFormer was able to achieve accurate spatial layout representation from invariant features extracted by the static encoder, minimizing the need for redundant coverage of an environment. Target hidden state concatenation enabled NavFormer to achieve improved implicit target detection during navigation by directly conditioning on the target hidden states. This aids the robot to recognize and reach the target object when it is within robot's camera view, thereby facilitating task completion. FCA had higher performance than NavFormer across all environments as it used additional information such as the map of the environment to select frontiers to explore, ground-truth state information of other robots to avoid collisions.

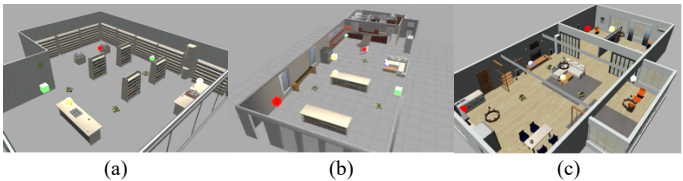

**Fig. 4.** Photorealistic environments with three jackal robots and eight target objects: (a) a bookstore, (b) a two-room cafe, and (c) a single floor house with multiple rooms.

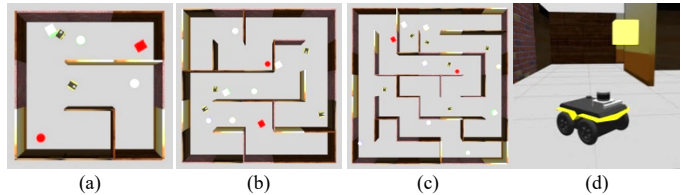

**Fig. 3.** Robot environments: (a) 7.5 m × 7.5 m with 2 robots, (b) 12.5 m × 12.5 m with 4 robots, (c) 17.5 m × 17.5 m with 6 robots, and (d) a mobile robot and a target object.

*B. Comparison Study in Unseen Photorealistic Environments*

We conducted experiments in three 3D photorealistic environments which included a bookstore (14.3 m × 15.6 m), a two-room cafe (9.3 m × 23.1 m), and a single-floor multi-room house (19.0 m × 11.0 m), as shown in Fig. 4. These environments represent unseen out-of-distribution environments with different configurations and appearances from our 3D maze environments used during training. The same target objects were used as in Section VII.A. Three Jackal robots were deployed in each environment. Both robot start and target object locations were randomly chosen. We performed 30 trials in each environment type for all methods.

*1) Results*: The results are presented in Table II. NavFormer achieved lower performance in unseen out-of-distribution photorealistic environments compared to its performance in unseen 3D mazes (Table I) as expected. This is similar to other visual tracking methods such as E-VAT [29]. However, NavFormer still outperformed VRW, DSAC, MA-TDN and DT in all three metrics. This is due to the data augmentation techniques utilized by NavFormer during the training of DVE, which allow for visual feature extraction to be robust to environmental changes. In turn, NavFormer has better generalization capabilities than these methods. As FCA uses additional sensory information about the environment and target, it is less sensitive to environmental variability.

*C. Comparison Study with Unseen Dynamic Obstacles*

We conducted simulated experiments with unseen dynamic obstacles to compare the robustness of the learned policy between NavFormer and other SOTA methods. During the dataset collection and training phases, only the Jackal robots were used as dynamic obstacles. To consider unseen dynamic obstacles during testing, we included mobile TurtleBot3 robots. The experiments were conducted in 3D mazes (12.5 m × 12.5 m), with a total of four robots deployed per trial: one Jackal robot, and three dynamic obstacles. Namely, we explored two scenarios, where: 1) all dynamic obstacles are Jackal robots, and 2) all dynamic obstacles are TurtleBot3 robots. Sixty trials were conducted for each scenario, with a new maze randomly generated every 10 trials. As in Section VII.A, we used the same target objects, however, both robot starting and target object locations were randomly chosen.

TABLE III
COMPARISON STUDY WITH SEEN AND UNSEEN DYNAMIC OBSTACLES

| | Metrics | VRW | DSAC | MA-TDN | DT | FCA | NavFormer |
|---|---|---|---|---|---|---|---|
| Seen Dynamic Obstacles | SR | 15.83% | 19.17% | 23.33% | 26.25% | 73.33% | 48.89% |
| | SPL | 0.104 | 0.133 | 0.137 | 0.201 | 0.482 | 0.378 |
| | CR | 0.680 | 0.460 | 0.394 | 0.430 | 0.096 | 0.223 |
| Unseen Dynamic Obstacles | SR | 13.33% | 16.67% | 21.67% | 21.67% | 75.00% | 41.67% |
| | SPL | 0.101 | 0.131 | 0.157 | 0.190 | 0.442 | 0.322 |
| | CR | 0.591 | 0.606 | 0.552 | 0.501 | 0.086 | 0.301 |

TABLE IV
ABLATION STUDY

| Methods | SR | SPL | CR |
|---|---|---|---|
| NavFormer | 48.89% | 0.378 | 0.223 |
| NavFormer w/o SSL | 39.58% | 0.272 | 0.376 |
| NavFormer w/o THSC | 32.08% | 0.261 | 0.500 |
| NavFormer w/o $\tau^e$ | 12.08% | 0.109 | 0.280 |
| NavFormer w/o $\tau^{ca}$ | 40.83% | 0.256 | 0.237 |
| NavFormer w/o $f_s$ | 41.25% | 0.289 | 0.230 |
| NavFormer w/o $f_g$ | 38.33% | 0.273 | 0.396 |
| NavFormer w ResNet18 | 7.92% | 0.080 | 0.765 |

*1) Results*: The results are shown in Table III. As expected, NavFormer's performance decreased slightly in the presence of unseen dynamic obstacles. However, NavFormer still outperformed the other SOTA methods, showcasing the robustness of its learned policy against other TDN SOTA methods for unseen dynamic obstacles. This is due to the CT network in NavFormer using the self-attention layers to dynamically adjust the weights of input robot historical trajectory that contain consecutive frames of visual observations. Thus, enabling the network to implicitly identify and adjust to the movement patterns of dynamic obstacles even when they have not been seen during training. FCA explicitly utilized ground-truth state information of other robots, therefore it achieved better performance than NavFormer.

*D. Ablation Study*

We conducted an ablation study to investigate the impact of the design choices on the training methods and architecture design of NavFormer. Namely, we considered:
(1) **NavFormer without SSL:** It was trained using only the loss of $\mathcal{L}^{DT}$, Eq. (10), (2) **NavFormer without Training on $\tau^e$:** When computing $\mathcal{L}^{DT}$, Eq. (10), robot exploration trajectories $\tau^e$ from $\mathcal{D}_e$ were not included, (3) **NavFormer without Training on $\tau^{ca}$:** When computing $\mathcal{L}^{DT}$, Eq. (10), robot collision avoidance trajectories $\tau^{ca}$ from $\mathcal{D}_{ca}$ were not included, (4) **NavFormer without Target Hidden States Concatenation (THSC):** The feed-forward layer of NavN includes input only from the hidden state of the observation $h_{o,t}^{ca}$ to generate $\hat{a}_t$, (5) **NavFormer without Static Encoder $f_s$:** We removed the static encoder during training, (6) **NavFormer without General Encoder $f_g$:** We removed the dynamic encoder during training, (7) **NavFormer with ResNet18:** We replaced the DVE with a single pre-trained ResNet18 [15].

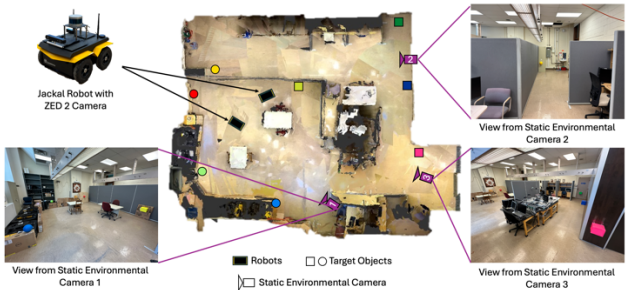

**Fig. 5.** Top view of the real-world office environment with two mobile robots and eight target objects with zoomed in views of the different regions.

We conducted 60 trials in 6 maze environments of 12.5 m × 12.5 m with 4 Jackal robots for each variant. A new environment is randomly generated every 10 trials.

*1) Results:* The results of the ablation study are presented in Table IV. NavFormer achieved the highest SR and SPL and the lowest CR among all variants. We found that NavFormer w/o $f_s$ achieved overall the second highest performance when compared to NavFormer, showing the significant role of the general encoder in extracting obstacle-specific features that are important for navigation. NavFormer with ResNet18 achieved the lowest performance among all variants due to the domain gap between the test environments (i.e., simulated 3D mazes) and ImageNet (i.e., web-scale object-centric images) [42]. The pretrained visual representations do not generalize to domain-specific robot navigation task, leading to the decrease in performance [43]. We also found that NavFormer w/o $\tau^{ca}$ achieved better performance than NavFormer w/o $\tau^e$. We postulate that this is due to the skill of exploration being more important for the completion of the TDN task. NavFormer inherently acquires the skill of collision avoidance with static obstacles during the learning of exploration skills, diminishing the incremental benefit of learning from $\tau^{ca}$.

VIII. REAL-WORLD EXPERIMENTS

We conducted real-world experiments in a 9.0 m by 10.0 m office environment, Fig. 5. Two Jackal robots were deployed with each having a ZED 2 camera for obtaining image observations. Eight target objects with varying geometric shapes (i.e., box, ball) and colors (i.e., blue, green, orange, yellow) were used. The same three datasets as described in Section V were collected: with 1) 50 single robot trajectories for exploration, 2) 100 two-robot trajectories for collision avoidance, and 3) 500 image pairs containing one image with dynamic obstacles and one image with only static obstacles. We finetuned NavFormer on these real-world datasets using the same set of hyperparameters as in Section VI and trained for 2,000 iterations. For testing, 15 trials were conducted with randomly placed target objects and robot starting locations.

NavFormer achieved an SR of 63.33%, SPL of 0.406, and CR of 0.229, respectively. This is comparable to the results in Table I for the small-mid-size environments. A video of our NavFormer method addressing the TDN problem in both the simulated and real-world environments is provided on our YouTube channel at https://youtu.be/PSVsLM1eGXo.

IX. CONCLUSION

In this paper, we present the development of a novel end-to-end DL model, NavFormer, to address the challenging problem of robot target-driven navigation in unknown and dynamic environments. Our approach uniquely combines a Dual-Visual

Encoder system with a transformer-based Navigation Network. The former is trained via self-supervised learning, while the latter uses offline reinforcement learning. Extensive simulated experiments were conducted with varying environment sizes and different number of dynamic obstacles. The results show that NavFormer outperformed state-of-the-art learning-based methods in successfully navigating to targets. Both photorealistic simulated experiments and real-world experiments validated the generalizability of NavFormer to effectively solve the TDN problem in dynamic unknown complex environments. An ablation study further validated our design choices of NavFormer. Future work will expand our architecture by incorporating object recognition backbones for common objects and different dynamic obstacles in everyday environments.

## REFERENCES

[1] A. H. Tan, F. P. Bejarano, Y. Zhu, R. Ren, and G. Nejat, "Deep Reinforcement Learning for Decentralized Multi-Robot Exploration With Macro Actions," *IEEE Robot. Autom. Lett.*, vol. 8, no. 1, pp. 272–279, Jan. 2023.

[2] A. Fung, B. Benhabib, and G. Nejat, "Robots Autonomously Detecting People: A Multimodal Deep Contrastive Learning Method Robust to Intraclass Variations," *IEEE Robot. Autom. Lett.*, vol. 8, no. 6, pp. 3550–3557, Jun. 2023.

[3] R. R. Murphy, "Activities of the rescue robots at the World Trade Center from 11-21 september 2001," *IEEE Robot. Autom. Mag.*, vol. 11, no. 3, pp. 50–61, Sep. 2004.

[4] A. H. Tan, S. Narasimhan, and G. Nejat, "4CNet: A Confidence-Aware, Contrastive, Conditional, Consistency Model for Robot Map Prediction in Multi-Robot Environments," *arXiv*:2402.17904, 2024.

[5] L. Mezghan et al., "Memory-Augmented Reinforcement Learning for Image-Goal Navigation," in *Proc. IEEE Int. Conf. Intell Robots Syst*, 2022, pp. 3316–3323.

[6] P. Fiorini and Z. Shiller, "Motion planning in dynamic environments using velocity obstacles," *Int. J. Robot. Res.*, vol. 17, no. 7, pp. 760–772, Jul. 1998.

[7] C. Yu et al., "DS-SLAM: A Semantic Visual SLAM towards Dynamic Environments," in *Proc. IEEE Int. Conf. Intell. Robot Syst.*, 2018, pp. 1168–1174.

[8] Y. Zhu et al., "Target-driven visual navigation in indoor scenes using deep reinforcement learning," in *Proc. IEEE Int. Conf. Robot. Aumomat.*, 2017, pp. 3357–3364.

[9] X. Ye, Z. Lin, H. Li, S. Zheng, and Y. Yang, "Active Object Perceiver: Recognition-guided Policy Learning for Object Searching on Mobile Robots," in *Proc. IEEE Int. Conf. Intell. Robots Syst.*, 2018, pp. 6857-6863.

[10] A. Devo, G. Mezzetti, G. Costante, M. L. Fravolini, and P. Valigi, "Towards Generalization in Target-Driven Visual Navigation by Using Deep Reinforcement Learning," *IEEE Trans. Robot.*, vol. 36, no. 5, pp. 1546–1561, Oct. 2020.

[11] W. Yang, X. Wang, A. Farhadi, A. Gupta, and R. Mottaghi, "Visual Semantic Navigation using Scene Priors," *arXiv*:1810.06543, 2018.

[12] X. Ye and Y. Yang, "Efficient Robotic Object Search Via HIEM: Hierarchical Policy Learning With Intrinsic-Extrinsic Modeling," *IEEE Robot. Autom. Lett.*, vol. 6, no. 3, pp. 4425–4432, Jul. 2021.

[13] Q. Wu, X. Gong, K. Xu, D. Manocha, J. Dong, and J. Wang, "Towards Target-Driven Visual Navigation in Indoor Scenes via Generative Imitation Learning," *IEEE Robot. Autom. Lett.*, vol. 6, no. 1, pp. 175–182, Jan. 2021.

[14] A. Radford, J. Wu, R. Child, D. Luan, D. Amodei, and I. Sutskever, "Language Models are Unsupervised Multitask Learners," *OpenAI blog*, 2019.

[15] K. He, X. Zhang, S. Ren, and J. Sun, "Deep Residual Learning for Image Recognition," in *Proc. IEEE Conf. Comput. Vis. Pattern Recognit.*, 2016, pp. 770–778.

[16] V. Mnih et al., "Asynchronous Methods for Deep Reinforcement Learning," in *Proc. Int. Conf. Mach. Learn.*, 2016, vol. 48, pp. 1928–1937.

[17] S. Hochreiter and J. Schmidhuber, "Long Short-Term Memory," *Neural Comput.*, vol. 9, no. 8, pp. 1735–1780, Nov. 1997.

[18] L. Espeholt et al., "IMPALA: Scalable Distributed Deep-RL with Importance Weighted Actor-Learner Architectures," in *Proc. Int. Conf. Mach. Learn.*, Jul. 2018, pp. 1407–1416.

[19] J. D. Van Berg, M. Lin, and D. Manocha, "Reciprocal velocity obstacles for real-time multi-agent navigation," in *Proc. IEEE Int. Conf. Robot. Automat.*, 2008, pp. 1928–1935.

[20] J. Alonso-Mora, A. Breitenmoser, M. Rufli, P. Beardsley, and R. Siegwart, "Optimal reciprocal collision avoidance for multiple non-holonomic robots," in *Distrib. Auton. Robot. Syst.*, Springer, 2013, pp. 203–216.

[21] P. Long, T. Fanl, X. Liao, W. Liu, H. Zhang, and J. Pan, "Towards optimally decentralized multi-robot collision avoidance via deep reinforcement learning," in *Proc. IEEE Int. Conf. Robot. and Automat.*, 2018, pp. 6252–6259.

[22] J. Zeng, R. Ju, L. Qin, Y. Hu, Q. Yin, and C. Hu, "Navigation in Unknown Dynamic Environments Based on Deep Reinforcement Learning," *Sensors*, vol. 19, no. 18, Art. no. 18, Jan. 2019.

[23] R. Han et al., "Reinforcement Learned Distributed Multi-Robot Navigation With Reciprocal Velocity Obstacle Shaped Rewards," *IEEE Robot. Autom. Lett.*, vol. 7, no. 3, pp. 5896–5903, Jul. 2022.

[24] M. Everett, Y. F. Chen, and J. P. How, "Motion Planning among Dynamic, Decision-Making Agents with Deep Reinforcement Learning," in *Proc. IEEE Int. Conf. Intell. Robots Syst.*, Dec. 2018, pp. 3052–3059.

[25] G. Sartoretti et al., "PRIMAL: Pathfinding via Reinforcement and Imitation Multi-Agent Learning," *IEEE Robot. Autom. Lett.*, vol. 4, no. 3, pp. 2378–2385, 2019.

[26] J. Schulman, F. Wolski, P. Dhariwal, A. Radford, and O. Klimov, "Proximal Policy Optimization Algorithms," pp. 1–12, 2017.

[27] K. Cho et al., "Learning Phrase Representations using RNN Encoder-Decoder for Statistical Machine Translation," 2014, *arXiv*:1406.1078.

[28] K.-H. Lee et al., "Multi-Game Decision Transformers," *Adv. Neural Inf. Process. Syst.*, pp. 27921-27936, 2022.

[29] A. Dionigi, A. Devo, L. Guiducci, and G. Costante, "E-VAT: An Asymmetric End-to-End Approach to Visual Active Exploration and Tracking," *IEEE Robot. Autom. Lett.*, vol. 7, no. 2, pp. 4259–4266, Apr. 2022.

[30] D. Fox, W. Burgard, and S. Thrun, "The dynamic window approach to collision avoidance," *IEEE Robot. Autom. Mag.*, vol. 4, no. 1, pp. 23–33, 1997.

[31] W. Khaksar, S. Vivekananthen, K. S. M. Saharia, M. Yousefi, and F. B. Ismail, "A review on mobile robots motion path planning in unknown environments," *Proc. IEEE Int. Symp. Robot. Intell. Sens.*, pp. 295–300, Apr. 2016.

[32] L. Chen et al., "Decision Transformer: Reinforcement Learning via Sequence Modeling," 2021, *arXiv*:2106.01345.

[33] P. Mirowski et al., "Learning to navigate in complex environments," *Proc. Int. Conf. Learn. Representations*, 2017.

[34] V. Mnih et al., "Human-level control through deep reinforcement learning," *Nature*, vol. 518, no. 7540, pp. 529–533, 2015.

[35] J.-B. Grill et al., "Bootstrap Your Own Latent - A New Approach to Self-Supervised Learning," in *Adv. Neural Inf. Process. Syst,* pp. 21271–21284, 2020.

[36] N. Srivastava, G. Hinton, A. Krizhevsky, I. Sutskever, and R. Salakhutdinov, "Dropout: A Simple Way to Prevent Neural Networks from Overfitting," *J. Mach. Learn. Res.*, pp. 1929–1958, 2014.

[37] R. S. Sutton and A. G. Barto, *Reinforcement learning: an introduction*, Second edition. in *Adaptive Comput. Mach. Learn. Ser.*, Cambridge, Massachusetts: The MIT Press, 2018.

[38] B. Yamauchi, "Frontier-based approach for autonomous exploration," in *Proc. IEEE Int. Symp. Compt. Intell. Robot. Automat.*, 1997.

[39] D. P. Kingma and J. Lei, "Adam: A Method for Stochastic Optimization," 2015, *arXiv*:1412.6980.

[40] I. Loshchilov and F. Hutter, "Decoupled Weight Decay Regularization," 2019, *arXiv*:1711.05101.

[41] P. Anderson et al., "On Evaluation of Embodied Navigation Agents," 2018, arXiv:1807.06757.

[42] J. Deng, W. Dong, R. Socher, L.-J. Li, Kai Li, and Li Fei-Fei, "ImageNet: A large-scale hierarchical image database," in *Proc. IEEE Conf. Comput. Vis. Pattern Recognit.,* 2009, pp. 248–255.

[43] A. Majumdar et al., "Where are we in the search for an Artificial Visual Cortex for Embodied Intelligence?" *arXiv*:2303.18240, 2024.